\documentclass[letter, 10pt, conference]{ieeeconf}      

\IEEEoverridecommandlockouts                              

\usepackage{cite}
\usepackage{amsmath,amssymb,amsfonts}
\usepackage[linesnumbered,ruled,vlined]{algorithm2e}
\usepackage{setspace}
\usepackage{bm}
\usepackage{graphicx}
\usepackage{textcomp}
\usepackage{xcolor}
\usepackage[scr=boondoxo,scrscaled=1.05]{mathalfa}
\usepackage{upgreek}

\def\BibTeX{{\rm B\kern-.05em{\sc i\kern-.025em b}\kern-.08em
    T\kern-.1667em\lower.7ex\hbox{E}\kern-.125emX}}

\newcommand{\R}{{\mathbb R}}

\newcommand{\be}{\begin{equation}}
\newcommand{\ee}{\end{equation}}
\newcommand{\ba}{\begin{array}}
\newcommand{\ea}{\end{array}}
\newcommand{\baa}{\left[\begin{array}}
\newcommand{\eaa}{\end{array}\right]}

\newcommand{\beqa}{\begin{eqnarray}}
\newcommand{\eeqa}{\end{eqnarray}}
\newcommand{\bt}{\begin{tabular}}
\newcommand{\et}{\end{tabular}}

\newcommand{\bi}{\begin{itemize}}
\newcommand{\ei}{\end{itemize}}
\newcommand{\bc}{\begin{center}}
\newcommand{\ec}{\end{center}}

\newtheorem{remark}{Remark}

\newcommand{\norm}[1]{\left\lVert#1\right\rVert}

\def\QED{\hfill \mbox{\rule[0pt]{1.5ex}{1.5ex}}}

\newcommand{\eor}{\ensuremath{\hfill\blacklozenge}}

\newlength{\algitab}
\begin{document}
\title{\LARGE \bf Data-driven Leak Localization in Water Distribution Networks via Dictionary Learning and Graph-based Interpolation
\thanks{Paul Irofti and Florin Stoican were supported by grants of the Romanian Ministry of Education and Research, CNCS - UEFISCDI,
project number PN-III-P1-1.1-PD-2019-0825
and
project number PN-III-P2-2.1-PED-2019-3248,
within PNCDI III.
Luis Romero and Vicenç Puig want to thank the Spanish national project DEOCS (DPI2016-76493-C3-3-R) and L-BEST (Ref. PID2020-115905RB-C21), as well as the Spanish State Research Agency through the María de Maeztu Seal of Excellence to IRI (MDM-2016-0656).}
}

\author{Paul Irofti$^1$, Luis Romero-Ben$^2$, Florin Stoican$^{1,3}$, Vicenç Puig$^2$ \thanks{$^1$ Research Center for Logic, Optimization and Security (LOS), Department of Computer Science, Faculty of Mathematics and Computer Science, University of Bucharest, Romania, {paul@irofti.net}}
\thanks{$^2$ Institut de Robòtica i Informàtica Industrial, Universitat Politècnica de Catalunya, Barcelona, Spain {\{luis.romero.ben,vicenc.puig\}@upc.edu}}
\thanks{$^3$ Politehnica University of Bucharest, Dept. of Automation Control and Systems Engineering, Romania, {florin.stoican@upb.ro}}}

\maketitle
\begin{abstract}
In this paper, we propose a data-driven leak localization method for water distribution networks (WDNs) which combines two complementary approaches: graph-based interpolation and dictionary classification. The former estimates the complete WDN hydraulic state (i.e., hydraulic heads) from real measurements at certain nodes and the network graph. Then, these actual measurements, together with a subset of valuable estimated states, are used to feed and train the dictionary learning scheme. 
Thus, the meshing of these two methods is explored, showing that its performance is superior to either approach alone, even deriving different mechanisms to increase its resilience to classical problems (e.g., dimensionality, interpolation errors, etc.). The approach is validated using the L-TOWN benchmark proposed at BattLeDIM2020.
\end{abstract}

\begin{keywords}
leak localization, dictionary learning, interpolation, water distribution networks.
\end{keywords}
\section{Introduction}

Fault detection and isolation (FDI) is an unavoidable element in any engineering system that is operated in an autonomous manner\cite{blanke2006diagnosis}. This has become especially relevant in the last decades due to the continued increase in complexity (in terms of size and interconnections) of networked systems. Water distribution networks (WDNs) are a prime example \cite{brdys1996operational}: they involve hundreds or even tens of thousands of nodes, have poor observability (traditionally outflows are measured only in tanks and reservoirs) and conservative control architecture (i.e., communication and decisions flow rigidly). All these issues make fault events (pipe bursts leading to leakage) hard to detect and isolate accurately.

Present day WDNs are becoming smarter by the addition of sensors, which provide timely information about pressures (node heads) and debits (pipe flows). This has made possible and raised considerable interest in two questions: 

\begin{itemize}
    \item how should the limited number of sensors be placed in the WDN such as to increase observability required for leak localization \cite{deshpande2013optimal,perelman2016sensor};
    \item based on measurements, what is the most efficient method to ensure leak detection and isolation \cite{perez2014leak,meseguer2014decision}.
\end{itemize}

Mainly, the difficulty of these questions comes from the problem size. 
Thus, there is a recent trend in the state of the art (for large-scale systems in general, but also for WDNs in particular \cite{taha2016methods}) to apply data-driven methods. This avoids the need of having an accurate mathematical model required by model-based approaches but neglects some physical relations that exist in the real system. 
A possible way to compensate the disregarding of these physical relations 
is by 
exploiting the system structure, i.e., the network's graph. 

This has led us to the idea of combining two recently developed and promising approaches: the graph-state interpolation (GSI) procedure proposed in \cite{Romero2021} and the dictionary learning (DL) procedure detailed in \cite{ISP20_jpc}. In this way, interpolated data (\textit{virtual sensors}) are added to the real measurements to feed the learning step of DL, hence indirectly considering network structural information during the leak localization process. Moreover, the use of a robust classification step, provided by DL, allows to improve the node-level classification performance of GSI (and its associated localization method in \cite{Romero2021}).

Additionally, a new application scheme is proposed to alleviate the drawbacks of introducing approximated data to the learning step of DL: separate dictionaries considering individual extra \textit{virtual sensors} (subsequently combined using a voting rule) are derived, instead of computing a single dictionary that encompasses all those \textit{virtual sensors}.



The theoretical elements are validated over the L-Town benchmark proposed in \cite{Battledim2020}, carrying out simulations on Epanet \cite{Epanet:2000} to generate the required fault scenarios.
\section{Preliminaries} \label{section:preliminaries}

We briefly introduce the theoretical background used in the rest of the paper. 
\subsection{Dictionary Learning Methods}\label{subsection:DL}

Recently,
we successfully applied dictionary learning techniques to solve WDN fault detection and isolation problems~\cite{IS17_ifac,ISP20_jpc}.
The DL problem trains an over-complete base $\bm{D}\in\R^{m\times n}$,
also called a dictionary,
on a dataset $\bm{Y}\in\R^{m\times N}$
to produce the $s$-sparse representations $\bm{X}\in\R^{n\times N}$:
\be
\begin{aligned}
& \underset{\bm{D}, \bm{X}}{\min}
& & \norm{\bm{Y}-\bm{DX}}_F^2 \\
& \text{s.t.}
& & \norm{\bm{x}_\ell}_{0} \leq s,\ \ell = 1:N \\
& & & \norm{\bm{d}_j} = 1, \ j = 1:n
\end{aligned}
\label{dict_learn}
\ee

The $s$ nonzero entries of a given dataset sample's representation
define the dictionary columns, also called atoms, and their associated coefficients.
A thorough description of the field is given in~\cite{DL_book}.

For water networks,
the dataset consists of sensor node pressure measurements
where each sample corresponds to a leak of a given magnitude
that occurs in one of the network nodes.
The FDI task consists of finding which node contains the leak. 
Looking at this as a classification problem,
where the network nodes represent the classes,
we can view
the dataset as multiple column blocks corresponding to
pressure measurements of leak scenarios occurring in a given node at various magnitudes.

For DL classification tasks,
Label Consistent K-SVD~(LC-KSVD)~\cite{JLD13} extends \eqref{dict_learn}
to simultaneously learn the linear classifier $\bm{W}$ based on labels $\bm{H}$
with an added discriminate constraint given by $\bm{Q}$
that enforces sets of atoms to only be used by a given class
\be 
\min_{\bm{D},\bm{W}, \bm{A}, \bm{X}} \|\bm{Y}-\bm{DX}\|_{F}^2 + \alpha \|\bm{H}-\bm{WX}\|_{F}^2
+ \beta \|\bm{Q}-\bm{AX}\|_{F}^2
\label{opt_label_cons}
\ee
Locating the leaky node from sensor measurements $\bm{y}$ is a two step process:
first we obtain the sparse representations $\bm{y}\approx\bm{Dx}$
by using a greedy algorithm such as OMP~\cite{PRK93omp};
secondly,
we perform linear classification $i=\arg\max\bm{Wx}$
to obtain the leaky node\footnote{Note that while $\bm Y$ contains only sensor node information, the labels in $\bm H$ help to identify leaky non-sensorized nodes.} $i$.
To accommodate large sets of data,
this approach was extended to the online semi-supervised setting in~\cite{IB19_toddler}, adapted and applied to large WDNs in~\cite{ISP20_jpc}.

\subsection{Interpolation strategies}\label{subsection:interpolation_strategies}


In the past, we have used interpolation methods as part of leak localization schemes \cite{Romero2021}. In particular, this interpolation process works over hydraulic head (node pressure + elevation) data, measured by pressure sensors (cheaper and easier to install than other metering systems \cite{Soldevila2021}). 

The network structure is considered to be represented by a graph $\mathcal{G}=(\mathcal{V},\mathcal{E})$, composed by a set of nodes, i.e., $\mathcal{V}=\mathcal R \cup \mathcal J$, that model the reservoirs ($\mathcal R$) and junctions ($\mathcal J$) of the WDN; and a set of edges, i.e., $\mathcal{E}$, representing the pipes of the network. The proposed interpolation procedure approximates the actual relation, given by the Hazen-Williams formula,  between the junctions' hydraulic heads using the following linear relation:

\begin{equation}\label{eq:1}
    f_i = \frac{1}{\phi_i}\bm{\omega}_i\bm{f}
\end{equation}

\noindent where $\bm{f}\in \mathbb{R}^{\lvert \mathcal{V}\rvert}$ is the complete graph-state vector, which encompasses the estimated and known hydraulic head values; $\bm{\omega}_i$ stands for the \textit{i-th} row of the weighted adjacency matrix $\bm{\Omega}\in \mathbb{R}^{\lvert \mathcal{V}\rvert \times \lvert \mathcal{V}\rvert}$, which encodes the connectivity among nodes as well as the strength of these connections; and $\phi_i=\sum_{j=1}^{\lvert \mathcal{V}\rvert} \omega_{ij}$ denotes the \textit{i-th} element of the diagonal of the degree matrix $\bm{\Phi}\in \mathbb{R}^{\lvert \mathcal{V}\rvert \times \lvert \mathcal{V}\rvert}$, which is a diagonal matrix. The weights at $\bm{\Omega}$ are derived from the actual pipe lengths: considering $p_{ij}$ to be the length of the pipe connecting nodes $\mathscr{v}_i$ and $\mathscr{v}_j$ ($p_{ij} = 0$ if they are not adjacent), then $\omega_{ij}=\frac{1}{p_{ij}}$ if $p_{ij}\neq 0$, and $\omega_{ij}=0$ otherwise. This selection increases the effect of closer neighbors over the state of a node.

Considering the previous approximation, the interpolation procedure can be expressed as a quadratic programming problem (see \cite{Romero2021} for the complete development):
\begin{subequations}
\label{eq:2}
\begin{align}
\label{eq:2_a}
\min_{\bm{f}} \quad & \frac{1}{2}\big[\bm{f}^T\bm{L}\bm{\Phi}^{-2}\bm{L}\bm{f}+\alpha \gamma ^2\big]\\
\label{eq:2_b}\textrm{s.t.} \quad & \bm{B}\bm{f}\leq \bm{1}_{\lvert \mathcal{V}\rvert}\cdot\gamma\\
\label{eq:2_c}& \gamma > 0 \\
\label{eq:2_d}&\bm{S}\bm{f}=\bm{f}^s  
\end{align}
\end{subequations}

\noindent where $\bm{L}$ stands for the graph Laplacian, $\bm{B}$ is the incidence matrix); $\bm{S}$ is a diagonal matrix with 1 at the elements whose associated node is sensorized (and 0 otherwise), $\bm{f}^s$ denotes the measurements vector, including the known hydraulic heads at the elements corresponding to metered nodes (and 0 elsewhere). $\alpha$  and $\gamma$ are auxiliary scalars which control the relative importance of the cost terms in \eqref{eq:2_a} and, respectively, relax the direction inequality constraint \eqref{eq:2_b}. 

The incidence matrix $\bm{B}$ deserves some additional explanation. Structural information (node position and elevation, pipe length) is usually available but information about flow direction within the pipes is much harder to get. A possible way to overcome this issue is to count edge crossing when enumerating the shortest paths from each of the reservoirs to each of the junction nodes. Subsequently, the flow along a pipe is taken in the direction with the higher count.

\begin{remark}
Enforcing the flow direction with $\bm B\bm f\leq 0$ may lead to an infeasible problem. We relax the inequality by the addition of slack variable $\gamma$ which is then penalized in the cost and weighted against the interpolation error via scalar $\alpha$. \eor
\end{remark}


\section{Methodology}
\label{sec:methodology}

The methodologies presented in Section \ref{section:preliminaries} have been successfully applied to solve the leak localization task, as illustrated in the provided references. Nevertheless, the performance of these techniques is limited by different aspects:

\begin{itemize}
    \item On the one hand, the DL approach introduced in Section \ref{subsection:DL} provides a satisfactory node-level localization, i.e., the leaks are mostly correctly located at the node where they appear. However, the method does not exploit the structure of the graph during the learning process, missing important information that may improve the performance.
    \item On the other hand, the GSI technique presented in Section \ref{subsection:interpolation_strategies}, together with the localization strategy proposed in \cite{Romero2021}, explicitly uses the graph structure for the sake of the leak isolation. However, the localization precision is lower, justified by the fact that the method was conceived to isolate an area where the leak is located rather than a specific node.
\end{itemize}

Thus, 
we propose a combination of GSI and DL that maximizes their benefits, leading to the presented methodology, henceforth referred as GSI-DL: 

\begin{enumerate}
    \item The hydraulic dataset (hydraulic heads at the sensorized nodes) is generated or provided, considering different leak scenarios (leak location, magnitude, occurring time, etc.); as well as leak-free historical information.
    \item The complete hydraulic state of the network (represented by the hydraulic heads at the WDN nodes) is estimated from the measured pressure values by means of GSI, for both the leak and leak-free scenarios, from which we derive the complete residuals dataset.
    \item A subset of nodes is selected to play the role of \textit{virtual sensors}, i.e., their interpolated state value is added to the information provided by the real sensors, obtaining an assembled residuals dataset.  
    \item The DL algorithm is fed with the assembled residuals dataset, performing the training and obtaining the corresponding dictionary and linear classifier.
\end{enumerate}
\begin{remark}
Step 3) deserves further explanations. The full set of interpolated values should not be used in the learning phase for two main reasons. First, the accumulation of differences between the actual hydraulic heads and the analogue interpolated states (produced by the approximations considered in GSI) reduce the leak localization accuracy. Second, the increase in computational complexity (storage and computation time) becomes significant, and it is no longer justified by the diminishing returns. \eor
\end{remark}
To conclude, the aim of the combined GSI-DL method is to merge the complementary strengths of each individual component (i.e., GSI supplies additional information for DL) in order to improve the classification performance. 

\subsection{Selection of the virtual nodes} \label{subsection:sel_virtual_nodes}

As previously mentioned, GSI retrieves the complete WDN state from a reduced set of known values from the physical sensors subset $\mathcal{S}_r \subset \mathcal{V}$. However, the introduced approximations produce differences between the actual hydraulic heads and the computed states at the unknown nodes, with an error distribution that strongly depends on the number and placement of sensors (to be expected, as they are the unique source of hydraulic data). 

Considering the importance of the existence of distinctive features for each leak, the insertion of estimated data must be carefully considered to avoid the inclusion of nodes whose values present high differences between actual and estimated value, between leaky and nominal data, etc., which can hinder the DL process and reduce the localization accuracy.


Thus, preliminary analyses must be performed by applying DL to datasets formed when considering the real measurements from $\mathcal{S}_r$ and different \textit{virtual sensors} subsets $\mathcal{S}_v$, in order to classify these combinations in terms of accuracy and select only those \textit{virtual sensors} adding valuable information.

\subsection{GSI-DL procedure} \label{subsection:GSI_DL}

To perform dictionary learning, the first step consists on collecting readings from pressure sensors in a wide variety of scenarios (as training samples are required): nominal/leaky, as well as different leak sizes and locations. However, due to the usual lack of this kind of data from the sensors of a real WDN, synthetic data is generated  by means of a well-calibrated hydraulic model.

Concretely, considering the typically large size of real networks, as well as the computational cost of performing simulations for every possible leak event, a subset $\mathcal{Z} \subseteq \mathcal{V}$ of nodes is chosen so that $\lvert \mathcal{Z} \rvert = c\leq N$ leak sources (labels for training) are considered, running scenarios with various fault magnitudes for each node $\mathscr{z}_i$. This subset of nodes must be scattered throughout the network to capture its hydraulic behaviour in the most complete way.


Noting that $M$ different leak sizes are considered for each possible leak location ($t$ time instants are computed considering each leak size), and that the number of physical sensors of the network is $\lvert \mathcal{S}_r \rvert$, the complete data set $\bm{\tilde{F}}_{leak}\in \R^{\lvert \mathcal{S}_r \rvert\times \psi} $, with $\psi=cMt$, may be regarded as the union of $c$ blocks corresponding to each faulty node $\mathscr{z}_i$, where each block contains $Mt$ samples with $M$ different leak sizes. 

Moreover, a complementary leak-free/nominal dataset $\bm{\tilde{F}}_{nom}\in \R^{\lvert \mathcal{S}_r \rvert\times \psi} $ is required: each column $\bm{\tilde{f}}_{nom}$ of this matrix must be obtained with similar boundary conditions at the network than in the case of its analogue (by position) column $\bm{\tilde{f}}_{leak}$ from $\bm{\tilde{F}}_{leak}$. In this way, the necessary residuals for DL are obtained while simultaneously reducing the effect of differences between leak and leak-free scenarios which are not caused by the leak.

The achieved datasets must be divided into their respective training and testing sets. Then, the learning process, explained in Section \ref{subsection:DL}, is applied as summarised by Algorithm \ref{alg:training} (consider that $s_{rv}=\lvert \mathcal{S}_r \rvert + \lvert \mathcal{S}_v \rvert$ is the total number of sensors). 

%
\begin{algorithm}
\SetKwComment{Comment}{}{}
\KwData{leak training set $\bm{\tilde{F}}_{leak}^{train} \in \R^{\lvert \mathcal{S}_r \rvert\times \psi_{tr}}$, nominal training set $\bm{\tilde{F}}_{nom}^{train} \in \R^{\lvert \mathcal{S}_r \rvert\times \psi_{tr}}$, sparsity level $s \in \R$, \textit{virtual sensors} $\mathcal{S}_v$, physical sensors $\mathcal{S}_r$ \\
}
\KwResult{dictionary~$\bm{D} \in \R^{s_{rv} \times n}$,~\mbox{classifier $\bm{W}\in\R^{c\times n}$}}
\setstretch{1.25}
interpolate $\bm{F}_{nom}^{train}$ applying (\ref{eq:2}) to $\bm{\tilde{F}}_{nom}^{train}$\\
interpolate $\bm{F}_{leak}^{train}$ applying (\ref{eq:2}) to $\bm{\tilde{F}}_{leak}^{train}$\\
compute residuals:   $\bm{Y}^{train}=\bm{F}_{nom}^{train}-\bm{F}_{leak}^{train}$\\
assemble dataset (and associated labels $\bm H_{rv}^{train}$): $\bm{Y}_{rv}^{train}=\big[\bm{Y}^{train}(\mathcal{S}_r);\bm{Y}^{train}(\mathcal{S}_v)\big]$\\
compute $\bm{D}, \bm{W}, \bm{A}$ applying (\ref{opt_label_cons}) to labeled $\bm{Y}_{rv}^{train}$

\caption{GSI-DL | Training procedure}
\label{alg:training}
\end{algorithm}

The obtained matrices, i.e., $\bm{D}$ and $\bm{W}$, are then used to classify the entries of the testing dataset in order to assess the reliability of the solution. Algorithm \ref{alg:toddler} summarizes the behavior of the classifier against new data entries.

\begin{algorithm}
\SetKwComment{Comment}{}{}
\KwData{sample $\bm{\tilde{f}}_{leak}^{test} \in \R^{\lvert \mathcal{S}_r \rvert}$, nominal sample $\bm{\tilde{f}}_{nom}^{test} \in \R^{\lvert \mathcal{S}_r \rvert}$, sparsity level $s \in \R$, \textit{virtual sensors} $\mathcal{S}_v$, physical sensors $\mathcal{S}_r$, dictionary $\bm{D} \in \R^{s_{rv} \times n}$, classifier $\bm{W}\in\R^{c\times n}$\\
}
\KwResult{faulty node $i$}
\setstretch{1.25}
interpolate $\bm{f}_{nom}^{test}$ applying (\ref{eq:2}) to $\bm{\tilde{f}}_{nom}^{test}$\\
interpolate $\bm{f}_{leak}^{test}$ applying (\ref{eq:2}) to $\bm{\tilde{f}}_{leak}^{test}$\\
compute residuals sample:   $\bm{y}^{test}=\bm{f}_{nom}^{test}-\bm{f}_{leak}^{test}$\\
assemble data: $\bm{y}_{rv}^{test}=\big[\bm{y}^{test}(\mathcal{S}_r);\bm{y}^{test}(\mathcal{S}_v)\big]$\\
representation: $\bm{x} = \text{OMP}(\bm{y}_{rv}^{test},\bm{D},s)$ \\
classification: $i = \arg\max_{j=1:c}(\bm{Wx})$ \\

\caption{GSI-DL | Classification procedure}
\label{alg:toddler}
\end{algorithm}


\subsection{GSI-DL with multiple dictionaries}\label{subsection:multiple_dictionaries}

The application and implementation of the previous strategy is limited due to the insertion of information from multiple interpolated nodes which leads to an accumulation of approximation errors and, above a threshold, deteriorates the efficacy of the leak localization process. Thus, a new scheme is conceived to overcome this problem: instead of learning a single dictionary/classifier that is trained with the entire set of additional virtual sensor information, we learn multiple dictionaries/classifiers, each one trained with a single additional \textit{virtual sensor}.

Then, to find the final classification result, a voting method \cite{Van2002} is applied to the set of partial classifications from the dictionary/classifier pairs. Several advantages are derived from the application of this scheme, chiefly, a more robust classification result and a reduction in the outliers' effect.


\section{Results}

\begin{figure}[!ht]
    \centering
    \includegraphics[width=\columnwidth]{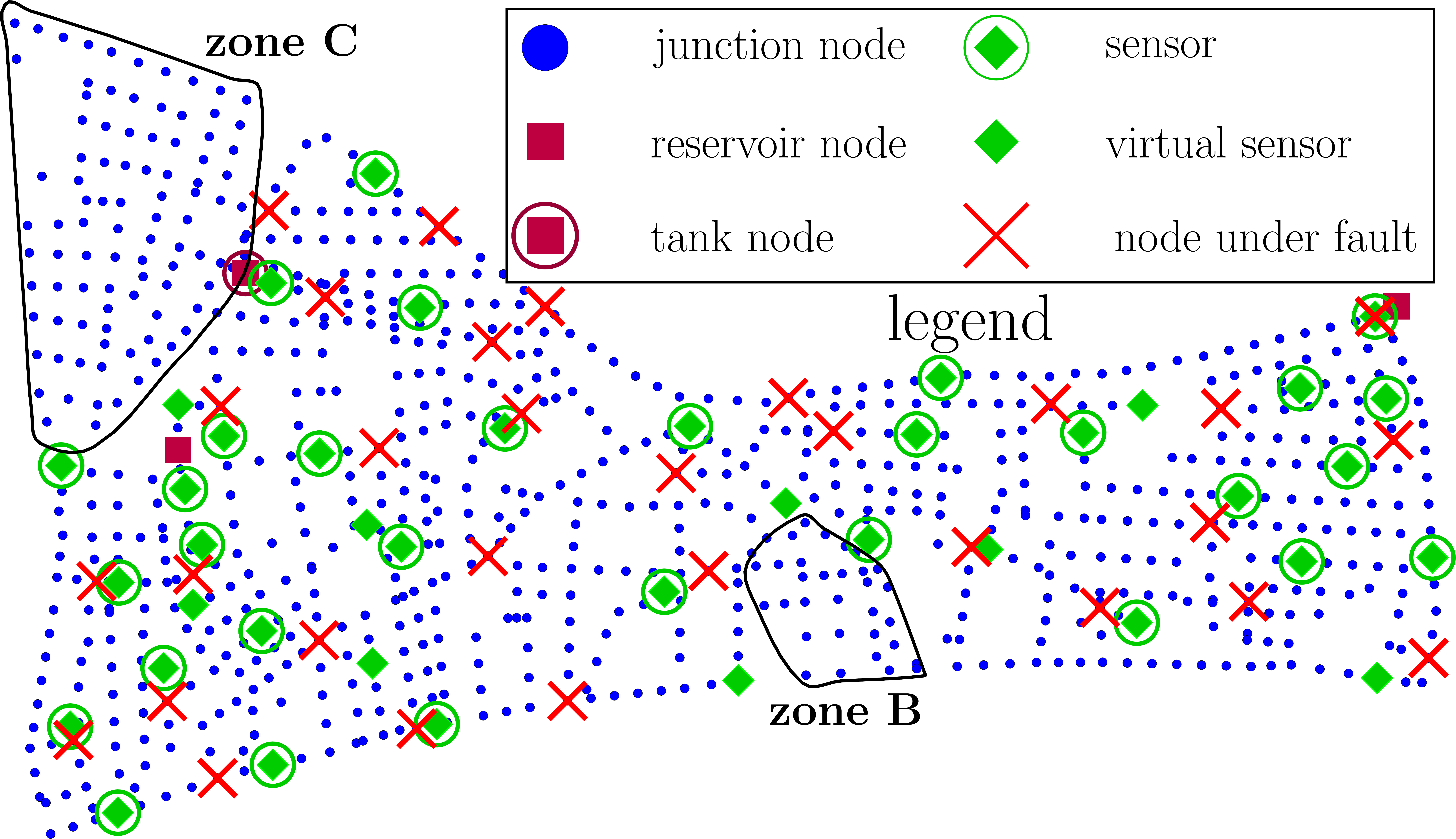}
    \caption{L-Town illustration with various node highlights}
    \label{fig:ltown}
\end{figure}

The presented methodologies are implemented in a realistic case study, provided by the ``Battle of Leakage Detection and Isolation Methods 2020'' (BattLeDIM2020), detailed in \cite{Battledim2020}. This benchmark, illustrated in Fig.\ref{fig:ltown},  consists of a small hypothetical network composed of 782 inner nodes, 909 pipes, 1 tank and 2 water inlets or reservoirs. The network is composed from three distinct areas (A, B and C), distinguished by the elevation of its nodes. 

Hereinafter, we focus on area A due to several reasons: 
\begin{itemize}
    \item the areas are separated by network elements, largely making them behave independently: areas A and B are connected by a pressure regulation valve (PRV); areas A and C are separated by a tank (filled from area A to feed area C);
    \item it is the larger area, composed of 659 junction nodes, and the one area directly connected to the water inlets;
    \item it has a high density of pressure sensors (29 of them, whereas there are only 3 in area C and one in area B).
\end{itemize}

\subsection{Data generation}

To test the methods proposed in Section~\ref{sec:methodology}, hydraulic data must be available for all the considered leaks. This information is retrieved, as further detailed in Section \ref{subsection:GSI_DL}, through the simulation of an EPANET \cite{Epanet:2000} model, part of the benchmark.

A list $\mathcal{Z}\subset \mathcal V$ of 30 nodes is selected from the junction nodes (denoted in Fig.\ref{fig:ltown} with ``filled blue circle'' symbols) as possible leak sources (``red cross'' symbols). Henceforth, 30 labels (one per each leak) are considered in the subsequent learning and classification steps. For each possible leak, a week of hydraulic data is generated by simulations, with a time step of 5 minutes. The first four days of the week are selected to generate the training set, increasing the leak size each day to cover a wide range of values, from 1 m\textsuperscript{3}/h to 7 m\textsuperscript{3}/h. The three remaining days produce the testing set, selecting even leak sizes between the minimum and maximum values to complete the set, i.e., 2, 4 and 6 m\textsuperscript{3}/h. Moreover, a 5\% of uncertainty is induced in the pipe roughness and diameters.

\subsection{Application of preliminaries}
\label{sec:preliminaries_application}

First, the preliminary works in Section \ref{section:preliminaries}, i.e., the separated application of dictionary learning \cite{ISP20_jpc} and GSI (together with a leak candidate selection method: GSI-LCSM) \cite{Romero2021} are applied to the aforementioned case study. This is done to provide performance baselines and to justify the interest in merging the procedures (i.e., to show the increase in performance for the combined method).

On the one hand, the classical DL approach is trained with the measurements from the real sensors. In this case, we obtain an accuracy of 89.63\%, which can be considered as the reference value to improve with the new proposed schemes that include the structural information from the interpolation. This accuracy is included in Table \ref{table:1} together with the rest of the single-dictionary results, as they are directly comparable.

On the other hand, GSI-LCSM is conceived to provide a candidate localization area, which limits the precision for single node localization. Thus, we need to select a suitable criteria to determine the success of the localization process, in order to perform a comparable analysis with respect to the dictionary learning ones. 

Let us consider that the GSI-LCSM method provides node $\mathscr{v}_{G}$ as the best candidate. If we only accept localization results as successful when the distance from $\mathscr{v}_{G}$ to $\mathscr{v}_{l}$ (node of the actual leak), i.e, $r(\mathscr{v}_{G},\mathscr{v}_{l})$; is the minimum among the distances from $\mathscr{v}_{G}$ to all the possible leaks, a reduced accuracy of 56.7\% is achieved. However, this criteria can be extended to also consider the case when $\mathscr{v}_{l}$ is not the closest to $\mathscr{v}_{G}$, but the 2\textsuperscript{nd}-nearest one (let us denote the closest one as $\mathscr{v}_{c}$). Accepting only the concrete cases when $r(\mathscr{v}_{G},\mathscr{v}_{l})-r(\mathscr{v}_{G},\mathscr{v}_{c})<100$ (meters), the accuracy is drastically increased, reaching a 76.7\%. Finally, an accuracy of 86.7\% can be achieved if we accept as successful all the cases when $\mathscr{v}_{l}$ is at least the 2\textsuperscript{nd}-nearest possible leak to $\mathscr{v}_{G}$.

Hence, similar conclusions to the ones stated in \cite{Romero2021} are extracted from these results: the GSI-LCSM method successfully reduces the uncertainty in the leak's position to a small zone around it, but the performance deteriorates if the localization degree is reduced to the node-level. This justifies the interest in combining GSI and DL, demonstrated to be suitable for node-level localization tasks.

\subsection{Single dictionary with virtual sensors}
\label{sec:single_DL}

We apply the methodology presented in Section~\ref{subsection:GSI_DL} to select a set of proper \textit{virtual nodes}, as per the criteria explained in Section~\ref{subsection:sel_virtual_nodes}. Thus, GSI is used over a leak dataset having $\lvert \mathcal{S}_r \rvert = 33$ measurements (from 29 junction nodes with pressure sensors, as well as 2 measuring devices monitoring the reservoirs' pressure, and other 2 known pressures at the output of the PRVs; depicted with a ``green circled diamond'' in Fig.~\ref{fig:ltown}), $c=30$ different leaks, $M=7$ different leak sizes and $t=288$ time slots (at every 5 minutes in 24 hours). 

The classification performance for a set of considered \textit{virtual sensors} is included in Table \ref{table:1}, whereas the location of those nodes within the network is represented in Fig.~\ref{fig:ltown} (denoted by ``green diamond'' markers). For each table entry, a different subset of \textit{virtual sensors} is selected from the interpolated ones proposed by GSI. Each accuracy value is averaged over five training runs, to reduce the effect of possible learning outliers. Comparing the first entry (the case without \textit{virtual sensors}) against the rest of the table, we conclude that:

\begin{table}[htbp]
\caption{Single dictionary - Accuracy results}
\begin{center}
\begin{tabular}{|c|c|}
\hline
\textbf{Virtual node/s}&\textbf{Testing accuracy (\%)} \\
\hline
None & 89.63 \\
\hline
255 & 90.54 \\
\hline
559 & 90.23 \\
\hline
51 & 90.43 \\
\hline
153 & 90.09 \\
\hline
504 & 90.57 \\
\hline
339 & 91.96 \\
\hline
265 & 89.77 \\
\hline
773 & 82.53 \\
\hline
214 & 81.34 \\
\hline
265, 255 & 89.42 \\
\hline
265, 255, 559 & 88.54 \\
\hline
\end{tabular}
\label{table:1}
\end{center}
\end{table}

\begin{itemize}
    \item generally, the addition of a single \textit{virtual sensor} (from node 255 to node 265 in the table) to the list of `real' ones increases the accuracy of the classification, and consequently, the leak localization performance;
    \item there exist nodes whose interpolated information degrades the performance (due to discrepancies between their nominal and leaky interpolated data), as seen, e.g., for nodes 773 and 214 in the table;
    \item lastly, the insertion of more than one interpolated value gradually lowers the accuracy, as shown in the last two entries of the table.
\end{itemize}

\begin{remark}
The numerical results of Table \ref{table:1} illustrate that while the combination of GSI-DL generally improves the performance, we cannot add randomly new virtual sensors and that, in any case their number is limited by possible inaccuracies from the interpolation step.\eor
\end{remark}

Worth mentioning is also the associated computational complexity. Increasing the number of sensors by adding virtual ones leads to larger computation times for the DL procedure. We show in Table~\ref{table:2} a time consumption analysis, carried for the same DL training parameters.

\begin{table}[htbp]
\caption{Single dictionary - Time consumption}
\begin{center}
\begin{tabular}{|c|c|}
\hline
\textbf{No. of added \textit{virtual sensors}}& \textbf{Time consumption (s)}\\
\hline
0 & 97.37 \\
\hline
193 & 119.01 \\
\hline
448 & 150.84 \\
\hline
628 (zone A + reservoirs) & 188.21 \\
\hline
\end{tabular}
\label{table:2}
\end{center}
\end{table}

\subsection{Multiple dictionaries and voting methods}

The performance degradation observed in Section~\ref{sec:single_DL}, associated to the inclusion of multiple \textit{virtual sensors}, justifies the development of the scheme introduced in Section \ref{subsection:multiple_dictionaries}. Thus, a set of multiple dictionary/linear classifier pairs is derived, considering a single \textit{virtual sensor} for each case. Moreover, a voting method is applied to achieve an improved classification result. 
In this case, a ``plurality voting'' scheme has been selected, i.e., the label with the maximum number of votes among the different dictionary/classifier pairs is selected as the definitive label for the classified sample. We apply the voting scheme to two different implementations, and present performance results in Table \ref{table:3}.:
\begin{itemize}
    \item (MD)DL: the classical DL construction is applied, with consecutive increase for the number of dictionaries;
    \item (MD)GSI-DL: each dictionary adds a single \textit{virtual sensor} (distinct from all the others inserted a priori), again, the number of dictionaries is augmented consecutively.
\end{itemize}

\begin{remark}\label{remark:voting}
The voting scheme deserves further explanations.
For both (MD)DL and (MD)GSI-DL,
the common thread is that multiple dictionaries and classifiers are generated (by varying the learning parameters) and then a decision is taken through a voting procedure.
For (MD)DL there is only one level of voting
(in between the dictionaries resulted from the learning parameter variation).
For (MD)GSI-DL we have two-level voting:
at the low-level, it is similar with the (MD)DL and
at the high-level, the voting is done by varying the set of sensors used to obtain the training dataset $\bm Y$;
each dataset is built from
a common set of $N_r$ physical sensors
with an additional virtual sensor
(out of the pre-selected candidates set $N_v$).\eor

\end{remark}

\begin{table}[htbp]
\caption{Multiple dictionaries - Accuracy results}
\begin{center}
\begin{tabular}{|c|c|c|}
\hline
 & \multicolumn{2}{|c|}{\textbf{Accuracy (\%)}} \\
\hline
\textbf{No. of dictionaries$^{\mathrm{*1,2}}$}& (MD)DL& (MD)GSI-DL \\
\hline
1$^{\mathrm{*3}}$ & 94.40 & 94.40 \\
\hline
2 & 93.91 & 93.57 \\
\hline
3 & 95.50 & 95.38 \\
\hline
4 & 95.17 & 95.78 \\
\hline
5 & 96.09 & 96.54 \\
\hline
6 & 95.78 & 96.36 \\
\hline
7 & 96.09 & 96.54 \\
\hline
\multicolumn{3}{p{\columnwidth}}{$^{\mathrm{*1}}$Considered \textit{virtual sensors}: 255, 559, 51, 153, 504, 339.}\\
\multicolumn{3}{p{\columnwidth}}{$^{\mathrm{*2}}$Number of dictionaries of different types (Remark \ref{remark:voting}).}\\
\multicolumn{3}{p{\columnwidth}}{$^{\mathrm{*3}}$The first dictionary is a classical DL for both cases.}
\end{tabular}
\label{table:3}
\end{center}
\end{table}




\subsection{Discussion}

Several conclusions may be derived from the results in Table \ref{table:3}, regarding the suitability of the proposed method.

Interestingly, the classification performance for both methods is almost identical for the first three entries (to be expected for the first entry since here a classical dictionary is computed in both cases) with significant improvements appearing only when considering four or more dictionaries in the voting step. This shows that subsequent dictionary/classifier pairs complement the classification performance of the standard DL, strengthening the common correct results and providing extra information to overcome possible weaknesses. 
Moreover, we note that (MD)GSI-DL surpasses (MD)DL when extra dictionary/classifier pairs with single \textit{virtual sensors} are included, confirming the appropriateness of this methodology over the single-dictionary approach.

\begin{figure}[htbp]
\centerline{\includegraphics[width=\linewidth]{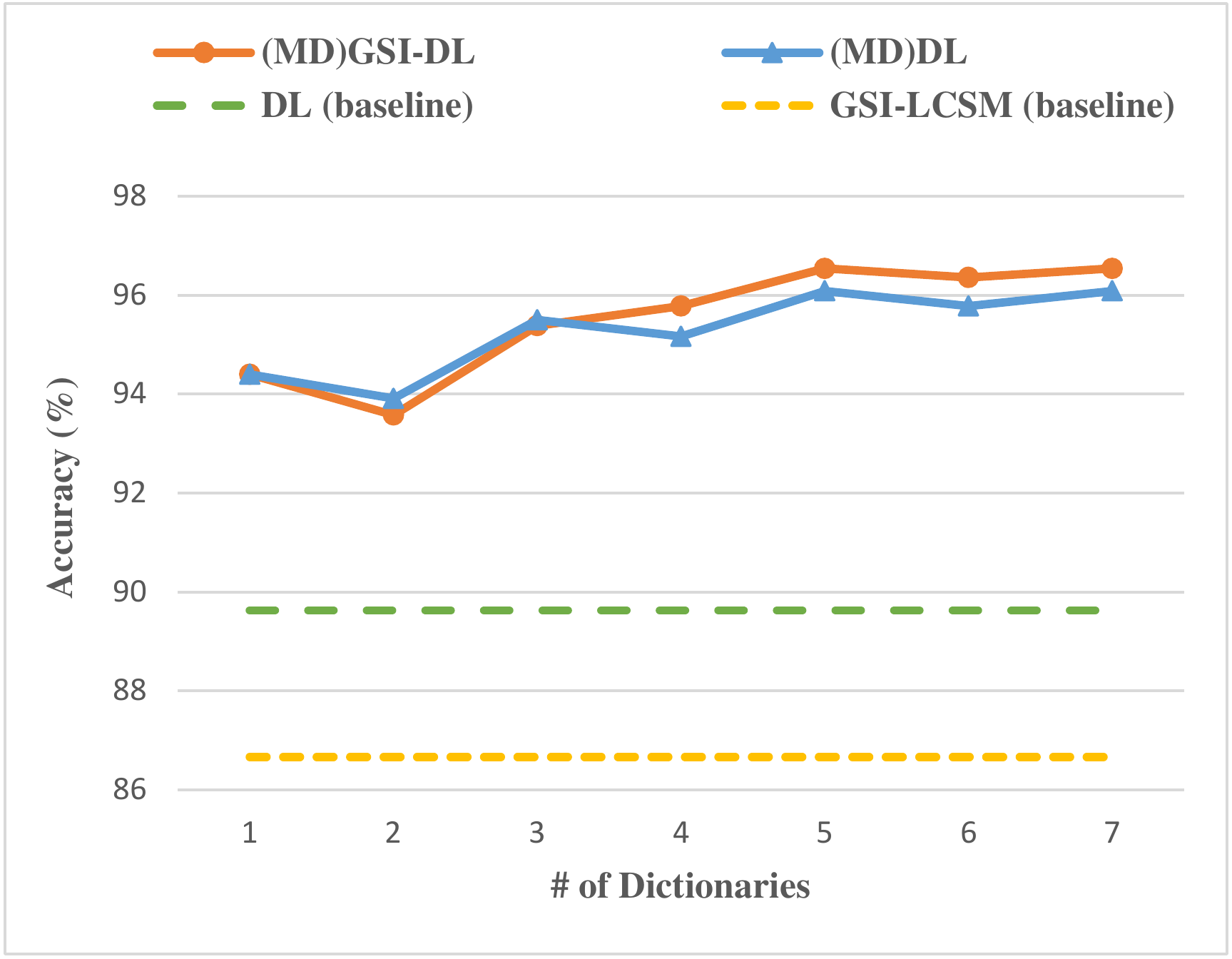}}
\caption{Performance comparison among the original and proposed methods}
\label{fig:3}
\end{figure}

Additionally, a comparison among the original and proposed approaches presented in this article, i.e., DL (with a single dictionary), GSI-LCSM, (MD)DL and (MD)GSI-DL, is represented in Fig.~\ref{fig:3}. The first two methods are used as baselines to assess the improvement of performance for the proposed methodologies. Thus, they are included in Fig.~\ref{fig:3} as dashed lines, despite that only one or even no dictionary is used their resolution. Additionally, for the proposed multiple dictionary approaches, the performance evolution regarding the number of ``different'' dictionaries (as mentioned above, three dictionaries of the same kind are used at each entry of Table \ref{table:3}) is included. 

This comparison illustrates the advantages of the multiple dictionary scheme, which increases the accuracy from 89.63\% to 94.4\% using the same base method (classical DL) with the only difference of deriving extra dictionaries and voting among them. Moreover, the improvement of (MD)GSI-DL with respect to (MD)DL is graphically represented to highlight the benefits of including \textit{virtual sensors} at the extra dictionaries.


\section{Conclusions}

This article proposes a methodology for leak localization in WDNs that combines graph-based state interpolation (GSI) and dictionary learning (DL). We have presented the original methods, explaining their characteristics, advantages and drawbacks, in order to show and justify the interest in their complementary application.

Furthermore, several different schemes have been introduced regarding this combination, i.e., computing a single dictionary that considers a set of interpolated values, deriving multiple dictionaries (classical or including \textit{virtual sensors}) and voting over their classification results, etc. 

The advantages and limitations of all these approaches have been stated and demonstrated by means of a case study, based on the L-Town benchmark from BattLeDIM2020. The achieved results show the performance improvements of the proposed methods.

Further works will cover additional enhancements of the methodology, such as a better integration of the components, an  analysis of the interpolation procedure to improve its performance (weight computation, nonlinear fitting, etc.).

\bibliographystyle{IEEEbib}
\bibliography{gsi-dl}      
\end{document}